# Some improved results on communication between information systems

Ping Zhu[a,b,*], Qiaoyan Wen[b]

[a]*School of Science, Beijing University of Posts and Telecommunications, Beijing 100876, China*
[b]*State Key Laboratory of Networking and Switching, Beijing University of Posts and Telecommunications, Beijing 100876, China*

**Abstract**

To study the communication between information systems, Wang et al. [C. Wang, C. Wu, D. Chen, Q. Hu, and C. Wu, Communicating between information systems, Information Sciences 178 (2008) 3228-3239] proposed two concepts of type-1 and type-2 consistent functions. Some properties of such functions and induced relation mappings have been investigated there. In this paper, we provide an improvement of the aforementioned work by disclosing the symmetric relationship between type-1 and type-2 consistent functions. We present more properties of consistent functions and induced relation mappings and improve upon several deficient assertions in the original work. In particular, we unify and extend type-1 and type-2 consistent functions into the so-called neighborhood-consistent functions. This provides a convenient means for studying the communication between information systems based on various neighborhoods.

## 1. Introduction

Rough set theory [7, 8] is a mathematical tool to deal with inexact or uncertain knowledge in information systems. It has originally described the indiscernibility of elements by equivalence relations. In order to handle incomplete information systems and complex practical problems, the requirement of equivalence relations has been relaxed to general binary relations in the literature (see [9, 10, 11, 14, 15, 16, 17, 18, 19, 22, 23] and the bibliographies therein) and the resultant systems are referred to as generalized approximation spaces by some authors. Although a great deal of work is concerned with investigating internally the structures and properties of a generalized approximation space, in recent years there are a few studies [1, 2, 3, 4, 13, 15, 16, 20] to compare the structures and properties of two generalized approximation spaces via homomorphisms or mappings. As explained in [12], homomorphisms allow one to translate the information contained in one granular world into the granularity of another granular world and thus provide a communication mechanism for exchanging information with other granular worlds.

In [16], Wang et al. introduced the notions of type-1 and type-2 consistent functions and investigated their properties from different perspectives. Moreover, they used relation mappings induced by mappings from generalized approximation spaces to arbitrary sets to construct relations on codomains. As an application of the properties of type-1 and type-2 consistent functions and relation mappings, they introduced two kinds of homomorphisms as a mechanism for communicating between information systems and gave some properties of information systems under homomorphisms. For example, they discussed some attributes of relation information systems by using type-1 and type-2 consistent functions and found out that the attribute reductions in the original system and its image system are equivalent under a certain condition. It turns out that consistent functions are useful for comparing the approximations and reductions in the original system and its image system. It should be pointed out that some other related works investigating information systems through homomorphisms [1, 2, 3, 4, 13, 20] are based on equivalence relations or other particular relations and are quite different from [15, 16].

The purpose of this paper is to provide an improvement of [16] by disclosing the symmetric relationship between type-1 consistent functions and type-2 consistent functions. We present more properties of consistent functions and induced relation mappings and improve upon several deficient assertions in [16]. More concretely, we unify and extend type-1 and type-2 consistent functions into the notion of neighborhood-consistent functions. We show that type-1 consistent functions are exactly predecessor-consistent functions, which reveals the symmetry of type-1 and type-2 consistent functions. Based on this observation, more properties of consistent functions are discovered. In addition, we greatly improve the theorem in [16] that describes the lower and upper approximations under relation mappings. We also present a new relationship between neighborhoods and relation mappings, which provides an approach to computing the predecessor and successor neighborhoods of an element of codomain with respect to the induced relation.

[*]Corresponding author
 *Email addresses:* `pzhubupt@gmail.com` (Ping Zhu), `wqy@bupt.edu.cn` (Qiaoyan Wen)

The remainder of this paper is structured as follows. In Section 2, we introduce the concept of neighborhood-consistent functions and show that type-1 consistent functions are exactly predecessor-consistent functions. Some properties of consistent functions are discussed in this section. Section 3 is devoted to amending and extending the properties of relation mapping, including the improvement of the theorem describing the lower and upper approximations under relation mappings in [16] and a new relationship between neighborhoods and relation mappings. We conclude the paper in Section 4.

## 2. Consistent functions

In this section, we examine consistent functions and their properties.

Let $U$ denote a finite and nonempty set called the universe. For each element $x$ of $U$, we may associate it with a subset $n(x)$ of $U$, called a *neighborhood* of $x$. Note that a neighborhood of $x$ may or may not contain $x$. The mapping $n : U \longrightarrow \mathscr{P}(U)$ is said to be a *neighborhood operator*, where we write $\mathscr{P}(U)$ for the power set of $U$. In addition, we follow generally used notation. In particular, the symbols $S_1 \backslash S_2$, $f(S)$, $f^{-1}(T)$ denote, respectively, the sets $\{x \mid x \in S_1, x \notin S_2\}$, $\{f(x) \mid x \in S\}$, and $\{x \in U \mid f(x) \in T\}$, where $f : U \longrightarrow V$ is a mapping, $S \subseteq U$, and $T \subseteq V$. With the notion of neighborhood, we can introduce the following definition.

**Definition 2.1.** Let $U$ and $V$ be finite and nonempty universal sets, and $n : U \longrightarrow \mathscr{P}(U)$ a neighborhood operator. A mapping $f : U \longrightarrow V$ is called a *neighborhood-consistent function* with respect to neighborhood operator $n$ if for any $x, y \in U$, $n(x) = n(y)$ whenever $f(x) = f(y)$.

Let $U$ be a finite and nonempty universal set, and suppose that $R \subseteq U \times U$ is a binary relation on $U$. For each $x \in U$, we associate it with a *predecessor neighborhood* $R_p(x)$ and a *successor neighborhood* $R_s(x)$ as follows [19]:

$$R_p(x) = \{y \in U \mid (y, x) \in R\}; \quad R_s(x) = \{y \in U \mid (x, y) \in R\}.$$

We call $R_p$ a *predecessor neighborhood operator* and call $R_s$ a *successor neighborhood operator*. More neighborhoods can be found in the literature [5, 6, 18, 19]. For example, based upon a binary relation $R$, one can define additional types of neighborhoods of $x$:

$$R_{p \wedge s}(x) = \{y \mid (x, y) \in R \text{ and } (y, x) \in R\} = R_p(x) \cap R_s(x),$$
$$R_{p \vee s}(x) = \{y \mid (x, y) \in R \text{ or } (y, x) \in R\} = R_p(x) \cup R_s(x).$$

In this paper, we are mainly concerned with the neighborhood operators $R_p$ and $R_s$. For later need, let us restate Definition 2.1 for the two operators.

**Definition 2.2.** Let $U$ and $V$ be finite and nonempty universal sets, $R$ a binary relation on $U$, and $f : U \longrightarrow V$ a mapping.

(1) The mapping $f$ is called a *predecessor-consistent function* with respect to $R$ if for any $x, y \in U$, $R_p(x) = R_p(y)$ whenever $f(x) = f(y)$.
(2) The mapping $f$ is called a *successor-consistent function* with respect to $R$ if for any $x, y \in U$, $R_s(x) = R_s(y)$ whenever $f(x) = f(y)$.

In other words, a mapping $f$ is predecessor-consistent (respectively, successor-consistent) if any two elements of $U$ with the same image under $f$ have the same predecessor (respectively, successor) neighborhood.

To illustrate the definition, let us see a simple example.

**Example 2.1.** Set $U = \{x_1, x_2, \ldots, x_7\}$ and $V = \{y_1, y_2, \ldots, y_6\}$. Take $R = \{(x_1, x_2), (x_1, x_3), (x_2, x_4), (x_3, x_4), (x_3, x_5), (x_4, x_6), (x_4, x_7), (x_5, x_6), (x_5, x_7)\}$. Define $f_i : U \longrightarrow V, i = 1, 2, 3$, as follows:

$$f_1(x_j) = y_j \text{ for } j = 1, 4, 5; f_1(x_2) = f_1(x_3) = y_2; f_1(x_6) = f_1(x_7) = y_6.$$
$$f_2(x_j) = y_j \text{ for } j = 1, 2, 3; f_2(x_4) = f_2(x_5) = y_4; f_2(x_6) = f_2(x_7) = y_6.$$
$$f_3(x_j) = y_j \text{ for } j = 1, 2, \ldots, 5; f_3(x_6) = f_3(x_7) = y_6.$$

Then by definition, it is easy to check that $f_1$ is predecessor-consistent (not successor-consistent) with respect to $R$, $f_2$ is successor-consistent (not predecessor-consistent) with respect to $R$, and $f_3$ is both predecessor-consistent and successor-consistent with respect to $R$.

We now recall the concept of consistent functions introduced in [16].

**Definition 2.3** ([16], Definition 3.1). Let $U$ and $V$ be finite universes, $R$ a binary relation on $U$, and $f : U \longrightarrow V$ a mapping. Let

$$[x]_f = \{y \in U \mid f(y) = f(x)\},$$
$$[x]_R = \{y \in U \mid R_s(y) = R_s(x)\}.$$

If $[x]_f \subseteq R_s(y)$ or $[x]_f \cap R_s(y) = \emptyset$ for any $x, y \in U$, then $f$ is called a *type-1 consistent function* with respect to $R$ on $U$. If $[x]_f \subseteq [x]_R$ for any $x \in U$, then $f$ is called a *type-2 consistent function* with respect to $R$ on $U$.

As we will see, the concept of type-1 (respectively, type-2) consistent function is equivalent to that of predecessor-consistent (respectively, successor-consistent) function in the sense of Definition 2.2. We prefer the latter term, as it is suggestive.

**Theorem 2.1.** *Let $U$ and $V$ be finite and nonempty universal sets and $R$ a binary relation on $U$.*

(1) *A mapping $f : U \longrightarrow V$ is a predecessor-consistent function with respect to $R$ if and only if it is a type-1 consistent function with respect to $R$.*
(2) *A mapping $f : U \longrightarrow V$ is a successor-consistent function with respect to $R$ if and only if it is a type-2 consistent function with respect to $R$.*

*Proof.* We need only to prove the first assertion; the second one follows directly from Definitions 2.2 and 2.3.

For (1), suppose that $f : U \longrightarrow V$ is a predecessor-consistent function with respect to $R$. In order to prove that $f$ is a type-1 consistent function, it suffices to show that for any $x, y \in U$, either $[x]_f \subseteq R_s(y)$ or $[x]_f \cap R_s(y) = \emptyset$. Equivalently, we need only to show that for any $x, y \in U$, if $[x]_f \cap R_s(y) \neq \emptyset$, then $[x]_f \subseteq R_s(y)$. Assume that $[x]_f \cap R_s(y) \neq \emptyset$. Then there exists $z \in [x]_f \cap R_s(y)$. Therefore, $(y, z) \in R$, namely, $y \in R_p(z)$. On the other hand, for any $w \in [x]_f$, we have that $f(w) = f(x) = f(z)$. By the definition of predecessor-consistent function, we see that $R_p(w) = R_p(z)$. Consequently, $y \in R_p(w)$, and thus $(y, w) \in R$, which means that $w \in R_s(y)$. As a result, we have that $[x]_f \subseteq R_s(y)$, and thus, $f$ is type-1 consistent with respect to $R$.

Conversely, assume that $f$ is a type-1 consistent function with respect to $R$. Suppose, by contradiction, that there exist $x_1, x_2 \in U$ with $f(x_1) = f(x_2)$ such that $R_p(x_1) \neq R_p(x_2)$. Without loss of generality, assume that there is $y \in R_p(x_1) \setminus R_p(x_2)$, that is, $(y, x_1) \in R$, while $(y, x_2) \notin R$. The former implies that $[x_1]_f \cap R_s(y) \neq \emptyset$ since $x_1 \in [x_1]_f \cap R_s(y)$, and the latter means that $[x_1]_f \nsubseteq R_s(y)$ because $x_2 \in [x_1]_f \setminus R_s(y)$. It contradicts with the definition of type-1 consistent function. Therefore, $f$ is predecessor-consistent by definition, finishing the proof of the theorem. □

Recall that a binary relation $R$ on $U$ is said to be *reflexive* if $(x, x) \in R$ for every $x \in U$; the relation $R$ is said to be *symmetric* if $(x, y) \in R$ implies $(y, x) \in R$ for any $x, y \in U$; the relation $R$ is said to be *transitive* if for any $x, y, z \in U$, $(x, y) \in R$ and $(y, z) \in R$ imply $(x, z) \in R$. For a binary relation $R$, the *inverse* $R^{-1}$ of $R$ is defined by

$$R^{-1} = \{(y, x) \mid (x, y) \in R\}.$$

Clearly, $R$ is reflexive (respectively, transitive) if and only if $R^{-1}$ is reflexive (respectively, transitive), and $R$ is symmetric if and only if $R = R^{-1}$. Observe that the predecessor neighborhood defined by $R$ is exactly the successor neighborhood defined by $R^{-1}$, and conversely, the successor neighborhood defined by $R$ is exactly the predecessor neighborhood defined by $R^{-1}$. Formally, for any $x \in U$,

$$R_p(x) = \{y \in U \mid (y, x) \in R\} = \{y \in U \mid (x, y) \in R^{-1}\} = R_s^{-1}(x), \quad (1)$$

$$R_s(x) = \{y \in U \mid (x, y) \in R\} = \{y \in U \mid (y, x) \in R^{-1}\} = R_p^{-1}(x). \quad (2)$$

Let $R$ and $Q$ be two binary relations on $U$. Defining $R \cup Q$ and $R \cap Q$ by set-theoretic union and intersection, respectively, we have the following equations [19]:

$$(R \cup Q)_p(x) = R_p(x) \cup Q_p(x), \quad (3)$$
$$(R \cup Q)_s(x) = R_s(x) \cup Q_s(x), \quad (4)$$
$$(R \cap Q)_p(x) = R_p(x) \cap Q_p(x), \quad (5)$$
$$(R \cap Q)_s(x) = R_s(x) \cap Q_s(x). \quad (6)$$

They follow directly from the definitions of predecessor and successor neighborhoods.

The following proposition clarifies the relationship between predecessor-consistent functions and successor-consistent functions. As a result, we may think that predecessor-consistent functions and successor-consistent functions are symmetric in some sense.

**Proposition 2.1.** *Let $U$ and $V$ be finite and nonempty universal sets and $R$ a binary relation on $U$.*

(1) *A mapping $f : U \longrightarrow V$ is predecessor-consistent with respect to $R$ if and only if it is successor-consistent with respect to $R^{-1}$.*

(2) *A mapping $f : U \longrightarrow V$ is successor-consistent with respect to $R$ if and only if it is predecessor-consistent with respect to $R^{-1}$.*

*Proof.* It follows immediately from Eqs. (1) and (2). □

**Corollary 2.1.** *Let $U$ and $V$ be finite and nonempty universal sets. If $R$ is a symmetric relation on $U$, then a mapping $f : U \longrightarrow V$ is predecessor-consistent with respect to $R$ if and only if it is successor-consistent with respect to $R$.*

*Proof.* It follows immediately from Proposition 2.1 and the fact that $R^{-1} = R$ if $R$ is symmetric. □

In addition, a predecessor-consistent function is exactly successor-consistent when $R$ is reflexive and transitive. To prove this, it is convenient to have the following lemma.

**Lemma 2.1.** *Let $R$ be a reflexive and transitive relation on $U$. Then for any $x, y \in U$, $R_p(x) = R_p(y)$ if and only if $R_s(x) = R_s(y)$.*

*Proof.* We only prove the necessity; the sufficiency can be verified similarly. By contradiction, assume that there exist $x, y \in U$ such that $R_p(x) = R_p(y)$, but $R_s(x) \neq R_s(y)$. Without loss of generality, suppose that there exists $z \in R_s(x) \setminus R_s(y)$. Then we see that $(x, z) \in R$, but $(y, z) \notin R$. Since $R$ is reflexive, we get that $y \in R_p(y) = R_p(x)$, namely, $(y, x) \in R$. We thus have by the transitivity of $R$ that $(y, z) \in R$, a contradiction. Consequently, $R_s(x) = R_s(y)$ and the necessity holds. □

Theorem 3.3 in [16] says that a mapping is a type-1 consistent function if and only if it is a type-2 consistent function, when the relation $R$ is reflexive and transitive. A simpler proof of this theorem arises from the above lemma.

**Theorem 2.2** ([16], Theorem 3.3). *Let $U$ and $V$ be finite and nonempty universal sets. If $R$ is a reflexive and transitive relation on $U$, then a mapping $f : U \longrightarrow V$ is predecessor-consistent with respect to $R$ if and only if it is successor-consistent with respect to $R$.*

*Proof.* It follows from Lemma 2.1. □

Suppose that $R$ and $Q$ are two binary relations on $U$, and $f : U \longrightarrow V$ is a mapping. Theorem 3.6 in [16] shows us that $f((R \cap Q)_s(x)) = f(R_s(x)) \cap f(Q_s(x))$ for any $x \in U$ if $f$ is predecessor-consistent with respect to both $R$ and $Q$. In fact, the next theorem shows us that this equation holds if $f$ is predecessor-consistent with respect to either $R$ or $Q$. By the way, we also present a similar property of successor-consistent functions.

**Theorem 2.3.** *Let R and Q be binary relations on U, and $f : U \longrightarrow V$ a mapping.*

(1) *If $f$ is predecessor-consistent with respect to either R or Q, then $f((R \cap Q)_s(x)) = f(R_s(x)) \cap f(Q_s(x))$ for any $x \in U$.*

(2) *If $f$ is successor-consistent with respect to either R or Q, then $f((R \cap Q)_p(x)) = f(R_p(x)) \cap f(Q_p(x))$ for any $x \in U$.*

*Proof.* (1) Without loss of generality, we may assume that $f$ is predecessor-consistent with respect to $R$. To prove $f((R \cap Q)_s(x)) = f(R_s(x)) \cap f(Q_s(x))$, it is sufficient to show that $f((R \cap Q)_s(x)) \supseteq f(R_s(x)) \cap f(Q_s(x))$ since the inverse inclusion is always true. For any $y \in f(R_s(x)) \cap f(Q_s(x))$, there are $z_1 \in R_s(x)$ and $z_2 \in Q_s(x)$ such that $f(z_1) = y = f(z_2)$. Therefore, $R_p(z_1) = R_p(z_2)$ by assumption. It follows from the fact $z_1 \in R_s(x)$ that $x \in R_p(z_1) = R_p(z_2)$, and thus $z_2 \in R_s(x)$. This, together with $z_2 \in Q_s(x)$, gives rise to $z_2 \in R_s(x) \cap Q_s(x) = (R \cap Q)_s(x)$. Consequently, $y = f(z_2) \in f((R \cap Q)_s(x))$, as desired.

(2) Again, without loss of generality, we may assume that $f$ is successor-consistent with respect to $R$. Whence, $f$ is predecessor-consistent with respect to $R^{-1}$ by Proposition 2.1. It follows from the first assertion and Eqs. (1), (2), (5), and (6) that

$$\begin{aligned} f((R \cap Q)_p(x)) &= f((R \cap Q)_s^{-1}(x)) \\ &= f((R^{-1} \cap Q^{-1})_s(x)) \\ &= f(R_s^{-1}(x)) \cap f(Q_s^{-1}(x)) \\ &= f(R_p(x)) \cap f(Q_p(x)), \end{aligned}$$

namely, $f((R \cap Q)_p(x)) = f(R_p(x)) \cap f(Q_p(x))$, finishing the proof of the theorem. □

For the union operation, any mapping preserves predecessor neighborhoods and successor neighborhoods.

**Proposition 2.2.** *Let R and Q be binary relations on U, and $f : U \longrightarrow V$ a mapping. Then for any $x \in U$,*

(1) $f((R \cup Q)_p(x)) = f(R_p(x)) \cup f(Q_p(x))$.
(2) $f((R \cup Q)_s(x)) = f(R_s(x)) \cup f(Q_s(x))$.

*Proof.* It follows directly from Eqs. (3) and (4). □

The next theorem complements Theorem 3.4 in [16], where the second part was missing.

**Theorem 2.4.** *Let R be a binary relation on U, and $f : U \longrightarrow V$ a mapping.*

(1) *The mapping $f$ is predecessor-consistent with respect to R if and only if $f^{-1}(f(R_s(x))) = R_s(x)$ for any $x \in U$.*

(2) *The mapping $f$ is successor-consistent with respect to R if and only if $f^{-1}(f(R_p(x))) = R_p(x)$ for any $x \in U$.*

*Proof.* (1) For the 'if' part, suppose, by contradiction, that there are $x_1, x_2 \in U$ with $f(x_1) = f(x_2)$ such that $R_p(x_1) \neq R_p(x_2)$. Without loss of generality, assume that there exists $z \in R_p(x_1) \setminus R_p(x_2)$. Therefore, $(z, x_1) \in R$ and $(z, x_2) \notin R$. The former means that $x_1 \in R_s(z)$. We thus get that $f(x_2) = f(x_1) \in f(R_s(z))$. Consequently, $x_2 \in f^{-1}(f(R_s(z))) = R_s(z)$, which forces that $(z, x_2) \in R$, a contradiction. Whence, $f$ is predecessor-consistent with respect to $R$

To see the 'only if' part, we may assume, again by contradiction, that there exists $z \in f^{-1}(f(R_s(x))) \setminus R_s(x)$ for some $x \in U$ because $R_s(x) \subseteq f^{-1}(f(R_s(x)))$ always holds. We thus get that $f(z) \in f(R_s(x))$. Hence, there is $y \in R_s(x)$ satisfying $f(y) = f(z)$. As $f$ is predecessor-consistent with respect to $R$, we see that $R_p(y) = R_p(z)$. This implies by the previous argument $y \in R_s(x)$ that $x \in R_p(z)$, namely, $z \in R_s(x)$, a contradiction. Thereby, $f^{-1}(f(R_s(x))) = R_s(x)$ for any $x \in U$.

(2) By Proposition 2.1, $f$ is successor-consistent with respect to $R$ if and only if it is predecessor-consistent with respect to $R^{-1}$. By the first assertion, this is equivalent to $f^{-1}(f(R_s^{-1}(x))) = R_s^{-1}(x)$, for any $x \in U$. Further, this is equivalent to $f^{-1}(f(R_p(x))) = R_p(x)$ for any $x \in U$, as $R_s^{-1}(x) = R_p(x)$. Thereby, (2) is true and this finishes the proof of the theorem. □

## 3. Relation mappings

In order to develop tools for studying the communication between two information systems, [16] explored relation mappings and their properties. This section is devoted to amending and extending some properties of relation mappings.

Let us review the definition of relation mappings introduced in [16].

**Definition 3.1** ([16], Definition 4.1). Let $f : U \longrightarrow V$ be a mapping. Then $f$ can induce a mapping $\hat{f} : \mathscr{P}(U \times U) \longrightarrow \mathscr{P}(V \times V)$ and a mapping $\hat{f}^{-1} : \mathscr{P}(V \times V) \longrightarrow \mathscr{P}(U \times U)$, that is,

$$\hat{f}(R) = \bigcup_{x \in U} \{f(x) \times f(R_s(x))\}, \text{ for any } R \in \mathscr{P}(U \times U),$$

$$\hat{f}^{-1}(Q) = \bigcup_{y \in V} \{f^{-1}(y) \times f^{-1}(Q_s(y))\}, \text{ for any } Q \in \mathscr{P}(V \times V).$$

We call $\hat{f}$ and $\hat{f}^{-1}$ *relation mapping* and *inverse relation mapping* induced by $f$, respectively.

The following is a compact, equivalent statement of Definition 3.1.

**Definition 3.2.** Let $U$ and $V$ be nonempty universal sets, and $f : U \longrightarrow V$ a mapping.

(1) The *relation mapping* induced by $f$, denoted by $\hat{f}$, is a mapping from $\mathscr{P}(U \times U)$ to $\mathscr{P}(V \times V)$ defined by

$$\hat{f}(R) = \{(f(x), f(y)) \mid (x, y) \in R\}$$

for all $R \in \mathscr{P}(U \times U)$.

(2) The *inverse relation mapping* induced by $f$, denoted by $\hat{f}^{-1}$, is a mapping from $\mathscr{P}(V \times V)$ to $\mathscr{P}(U \times U)$ defined by

$$\hat{f}^{-1}(Q) = \{(x, y) \in U \times U \mid (f(x), f(y)) \in Q\}$$

for all $Q \in \mathscr{P}(V \times V)$.

To illustrate the above definition, let us revisit Example 2.1.

**Example 3.1.** Recall that in Example 2.1, $U = \{x_1, x_2, \ldots, x_7\}$, $V = \{y_1, y_2, \ldots, y_6\}$, and $R = \{(x_1, x_2), (x_1, x_3), (x_2, x_4), (x_3, x_4), (x_3, x_5), (x_4, x_6), (x_4, x_7), (x_5, x_6), (x_5, x_7)\} \in \mathscr{P}(U \times U)$. Consider $f_1 : U \longrightarrow V$ defined by

$$f_1(x_j) = y_j \text{ for } j = 1, 4, 5; f_1(x_2) = f_1(x_3) = y_2; f_1(x_6) = f_1(x_7) = y_6.$$

Then it follows by definition that

$$\hat{f}_1(R) = \{(y_1, y_2), (y_2, y_4), (y_2, y_5), (y_4, y_6), (y_5, y_6)\},$$
$$\hat{f}_1^{-1}(\hat{f}_1(R)) = \{(x_1, x_2), (x_1, x_3), (x_2, x_4), (x_2, x_5), (x_3, x_4), (x_3, x_5), (x_4, x_6), (x_4, x_7), (x_5, x_6), (x_5, x_7)\}.$$

Recall that in [16], Theorem 4.2 (4) says that when the mapping $f : U \longrightarrow V$ is surjective and predecessor-consistent with respect to $R \subseteq U \times U$, the transitivity of $R$ implies that of $\hat{f}(R)$. In fact, the requirement that $f$ is surjective is not necessary. Moreover, we find that the successor-consistent function has the same property.

**Theorem 3.1.** *Suppose that $R \subseteq U \times U$ is transitive and $f : U \longrightarrow V$ is successor-consistent with respect to $R$. Then $\hat{f}(R)$ is transitive.*

*Proof.* For any $(y_1, y_2) \in \hat{f}(R)$ and $(y_2, y_3) \in \hat{f}(R)$, there exist $(x_1, x_2) \in R$ and $(x_2', x_3) \in R$ satisfying $f(x_1) = y_1$, $f(x_2) = f(x_2') = y_2$, and $f(x_3) = y_3$. Therefore, we see that $x_3 \in R_s(x_2') = R_s(x_2)$, which means that $(x_2, x_3) \in R$. It follows from the transitivity of $R$ that $(x_1, x_3) \in R$, and thus,

$$(y_1, y_3) = (f(x_1), f(x_3)) \in \hat{f}(R).$$

This proves the transitivity of $\hat{f}(R)$. □

Let $f : U \longrightarrow V$ be a mapping, and $R, Q \in \mathscr{P}(U \times U)$. In [16], Theorem 4.3 (3) says that if $f$ is predecessor-consistent and successor-consistent with respect to both $R$ and $Q$, then $\hat{f}(R \cap Q) = \hat{f}(R) \cap \hat{f}(Q)$. In fact, the requirement of $f$ can be relaxed as follows.

**Theorem 3.2.** *Let $f : U \longrightarrow V$ be a mapping, and $R, Q \in \mathscr{P}(U \times U)$. Then $\hat{f}(R \cap Q) = \hat{f}(R) \cap \hat{f}(Q)$ if one of the following conditions holds.*

(1) *The mapping $f$ is both predecessor-consistent and successor-consistent with respect to $R$.*
(2) *The mapping $f$ is both predecessor-consistent and successor-consistent with respect to $Q$.*
(3) *The mapping $f$ is predecessor-consistent with respect to $R$ and successor-consistent with respect to $Q$.*
(4) *The mapping $f$ is successor-consistent with respect to $R$ and predecessor-consistent with respect to $Q$.*

*Proof.* We only prove (1) and (3), because of the symmetry of the assertions. Note that $\hat{f}(R \cap Q) \subseteq \hat{f}(R) \cap \hat{f}(Q)$ always holds by definition. Hence, we need only to verify the inverse inclusion. For any $(z_1, z_2) \in \hat{f}(R) \cap \hat{f}(Q)$, there exist $(x_1, x_2) \in R$ and $(y_1, y_2) \in Q$ such that $f(x_1) = f(y_1) = z_1$ and $f(x_2) = f(y_2) = z_2$. It remains to check that $(z_1, z_2) \in \hat{f}(R \cap Q)$.

Let us begin with (1). Since $f$ is both predecessor-consistent and successor-consistent with respect to $R$, we have that $R_s(x_1) = R_s(y_1)$ and $R_p(x_2) = R_p(y_2)$. It follows from $(x_1, x_2) \in R$ that $x_2 \in R_s(x_1) = R_s(y_1)$, and thus, $y_1 \in R_p(x_2) = R_p(y_2)$, namely $(y_1, y_2) \in R$. Combining this with the fact that $(y_1, y_2) \in Q$, we get that $(y_1, y_2) \in R \cap Q$, and thus, $(z_1, z_2) = (f(y_1), f(y_2)) \in \hat{f}(R \cap Q)$. Therefore, we get that $\hat{f}(R) \cap \hat{f}(Q) \subseteq \hat{f}(R \cap Q)$, as desired.

For (3), because $f$ is predecessor-consistent with respect to $R$ and successor-consistent with respect to $Q$, we obtain that $R_p(x_2) = R_p(y_2)$ and $Q_s(x_1) = Q_s(y_1)$. The former gives rise to $x_1 \in R_p(x_2) = R_p(y_2)$, namely, $(x_1, y_2) \in R$, while the latter yields that $y_2 \in Q_s(y_1) = Q_s(x_1)$, i.e., $(x_1, y_2) \in Q$. We thus get that $(x_1, y_2) \in R \cap Q$, which implies that $(z_1, z_2) = (f(x_1), f(y_2)) \in \hat{f}(R \cap Q)$. Therefore, $\hat{f}(R) \cap \hat{f}(Q) \subseteq \hat{f}(R \cap Q)$, finishing the proof of the theorem. □

The next theorem extends the assertion (2) of Theorem 4.6 in [16], where only the sufficiency has been provided.

**Theorem 3.3.** *Let $f : U \longrightarrow V$ be a mapping and $R \subseteq U \times U$. Then $\hat{f}^{-1}(\hat{f}(R)) = R$ if and only if $f$ is both predecessor-consistent and successor-consistent with respect to $R$.*

*Proof.* We first prove the necessity. Assume, by contradiction, that $f$ is not predecessor-consistent. Then there are $x_1, x_2 \in U$ with $f(x_1) = f(x_2)$ such that $R_p(x_1) \neq R_p(x_2)$, say, $z \in R_p(x_1) \setminus R_p(x_2)$. That is, $(z, x_1) \in R$ and $(z, x_2) \notin R$. We thus find that

$$(f(z), f(x_2)) = (f(z), f(x_1)) \in \hat{f}(R).$$

Hence,

$$(z, x_2) \in \hat{f}^{-1}(\hat{f}(R)) = R,$$

namely, $(z, x_2) \notin R$, which is absurd. Consequently, $f$ is predecessor-consistent with respect to $R$. Similarly, it is easy to show that $f$ is also successor-consistent with respect to $R$. Whence, the necessity is true.

One may refer to [16] for the proof of the sufficiency. For the convenience of the reader, we give another proof in our context. It is obvious that $\hat{f}^{-1}(\hat{f}(R)) \supseteq R$. Let us verify that $\hat{f}^{-1}(\hat{f}(R)) \subseteq R$. For any $(y_1, y_2) \in \hat{f}^{-1}(\hat{f}(R))$, we have by the definition of inverse relation mapping that $(f(y_1), f(y_2)) \in \hat{f}(R)$. Therefore, there is $(x_1, x_2) \in R$ such that $f(x_1) = f(y_1)$ and $f(x_2) = f(y_2)$. Since $f$ is both predecessor-consistent and successor-consistent with respect to $R$, we get that $R_s(x_1) = R_s(y_1)$ and $R_p(x_2) = R_p(y_2)$. It follows that $x_1 \in R_p(x_2) = R_p(y_2)$, which implies that $y_2 \in R_s(x_1) = R_s(y_1)$, namely, $(y_1, y_2) \in R$. As a result, we have that $\hat{f}^{-1}(\hat{f}(R)) = R$. This completes the proof of the theorem. □

Now, we would like to establish a relationship between neighborhoods and relation mappings.

**Theorem 3.4.** *Let $f : U \longrightarrow V$ be a mapping and $R \subseteq U \times U$. Then for any $x \in U$,*

(1) $\hat{f}(R)_p(f(x)) = \bigcup\limits_{f(x')=f(x)} f(R_p(x'))$. *In particular, $\hat{f}(R)_p(f(x)) = f(R_p(x))$ if $f$ is predecessor-consistent with respect to $R$.*

(2) $\hat{f}(R)_s(f(x)) = \bigcup\limits_{f(x')=f(x)} f(R_s(x'))$. *In particular, $\hat{f}(R)_s(f(x)) = f(R_s(x))$ if $f$ is successor-consistent with respect to $R$.*

*Proof.* We only prove the first assertion, since the second one can be proved similarly. For any $z \in \hat{f}(R)_p(f(x))$, there is $(z', x') \in R$ such that $f(z') = z$ and $f(x') = f(x)$. Therefore, $z' \in R_p(x')$, which implies that $z = f(z') \in f(R_p(x'))$. Hence,
$$z \in \bigcup_{f(x')=f(x)} f(R_p(x')).$$

This means that
$$\hat{f}(R)_p(f(x)) \subseteq \bigcup_{f(x')=f(x)} f(R_p(x')).$$

Conversely, for any
$$z \in \bigcup_{f(x')=f(x)} f(R_p(x')),$$

there is $x' \in U$ with $f(x') = f(x)$ such that $z \in f(R_p(x'))$. Whence, there exists $y \in R_p(x')$ satisfying $f(y) = z$. It follows that $(y, x') \in R$, and thus, $(f(y), f(x')) \in \hat{f}(R)$. Thanks to $f(x') = f(x)$, it yields that $(f(y), f(x)) \in \hat{f}(R)$, that is, $z = f(y) \in \hat{f}(R)_p(f(x))$. Consequently,
$$\bigcup_{f(x')=f(x)} f(R_p(x')) \subseteq \hat{f}(R)_p(f(x)),$$

and thus,
$$\hat{f}(R)_p(f(x)) = \bigcup_{f(x')=f(x)} f(R_p(x')),$$

as desired.

If $f$ is predecessor-consistent with respect to $R$, then for any $x' \in U$ with $f(x') = f(x)$, we have by definition that $R_p(x') = R_p(x)$. This gives rise to
$$\hat{f}(R)_p(f(x)) = \bigcup_{f(x')=f(x)} f(R_p(x')) = f(R_p(x)).$$

Hence, the first assertion holds. □

**Remark 3.1.** Note that Theorem 3.4 provides an approach to computing the predecessor and successor neighborhoods of an element of $V$ with respect to $\hat{f}(R)$. In fact, for any $y \in V$, if $y \notin f(U)$, then it is clear that $\hat{f}(R)_p(y) = \hat{f}(R)_s(y) = \emptyset$. Otherwise, there is some $x \in U$ such that $f(x) = y$, and thus, one may use Theorem 3.4 to compute $\hat{f}(R)_p(y)$ and $\hat{f}(R)_s(y)$.

To state the next theorem, we need to recall the notion of approximations. Let $U$ be a finite and nonempty universal set, and let $R \subseteq U \times U$ be a binary relation on $U$. The ordered pair $(U, R)$ is referred to as a *generalized approximation space*. For any $X \subseteq U$, one can characterize $X$ by a pair of lower and upper approximations (see, for example, [18, 19]). The *lower approximation* $\underline{apr}_R X$ and *upper approximation* $\overline{apr}_R X$ of $X$ are defined as
$$\underline{apr}_R X = \{x \in U \mid R_s(x) \subseteq X\} \text{ and } \overline{apr}_R X = \{x \in U \mid R_s(x) \cap X \neq \emptyset\},$$

respectively.

In [16], Theorem 4.8 (1-6) investigate the lower and upper approximations under relation mappings. For the sake of comparison, let us review the results.

**Theorem 3.5** ([16], Theorem 4.8). *Let $f : U \longrightarrow V$ be a mapping and $R \subseteq U \times U$.*

(1) *If $f$ is successor-consistent with respect to $R$, then*
$$f(\underline{apr}_R X) \subseteq \underline{apr}_{\hat{f}(R)} f(X)$$

*for any $X \subseteq U$.*

(2) *If f is both predecessor-consistent and successor-consistent with respect to R, then*

$$f(\underline{apr}_R X) = \underline{apr}_{\hat{f}(R)} f(X) = f(X)$$

*for any R-definable set $X \subseteq U$.*

(3) *If f is bijective, then*

$$f(\underline{apr}_R X) = \underline{apr}_{\hat{f}(R)} f(X)$$

*for any $X \subseteq U$.*

(4) *If f is successor-consistent with respect to R, then*

$$f(\overline{apr}_R X) \supseteq \overline{apr}_{\hat{f}(R)} f(X)$$

*for any $X \subseteq U$.*

(5) *If f is both predecessor-consistent and successor-consistent with respect to R, then*

$$f(\overline{apr}_R X) = \overline{apr}_{\hat{f}(R)} f(X) = f(X)$$

*for any R-definable set $X \subseteq U$.*

(6) *If f is bijective, then*

$$f(\overline{apr}_R X) = \overline{apr}_{\hat{f}(R)} f(X)$$

*for any $X \subseteq U$.*

**Remark 3.2.** Let us remark that the assertions (2) and (4) do not hold in general. For (2), consider the case that $f$ is not surjective. Then for any $y \in V \setminus f(U)$, we have that $\hat{f}(R)_s(y) = \emptyset \subseteq f(X)$ and thus

$$y \in \underline{apr}_{\hat{f}(R)} f(X).$$

But

$$y \notin f(\underline{apr}_R X) \subseteq f(U)$$

since $y \notin f(U)$. Hence,

$$f(\underline{apr}_R X) = \underline{apr}_{\hat{f}(R)} f(X) = f(X)$$

is not true in this case.

For (4), let us consider a counter example. Take $U = \{x, y, z\}$, $V = \{a, b\}$, and $R = \{(x, y)\}$, and define $f$ as follows:

$$f(x) = a, \quad f(y) = f(z) = b.$$

Clearly, $\hat{f}(R) = \{(a, b)\}$, and moreover, $f$ is successor-consistent with respect to $R$. Taking $X = \{z\}$, we find that $\overline{apr}_R X = \emptyset$ and thus $f(\overline{apr}_R X) = \emptyset$. On the other hand, we have that

$$f(X) = \{b\} \text{ and } \overline{apr}_{\hat{f}(R)} f(X) = \{a\}.$$

Therefore,

$$f(\overline{apr}_R X) \not\supseteq \overline{apr}_{\hat{f}(R)} f(X),$$

and the assertion (4) in Theorem 3.5 is false.

Let us present an improved version of Theorem 3.5.

**Theorem 3.6.** *Let $f : U \longrightarrow V$ be a mapping and $R \subseteq U \times U$.*

(1) *If f is successor-consistent with respect to R, then*

$$f(\underline{apr}_R X) \subseteq \underline{apr}_{\hat{f}(R)} f(X)$$

*for any $X \subseteq U$.*

(2) *If f is surjective, then*

$$\underline{apr}_{\hat{f}(R)} f(X) \subseteq f(\underline{apr}_R X)$$

*for any $X \subseteq U$ with $f^{-1}(f(X)) = X$.*

(3) *If f is surjective and successor-consistent with respect to R, then*

$$f(\underline{apr}_R X) = \underline{apr}_{\hat{f}(R)} f(X)$$

*for any $X \subseteq U$ with $f^{-1}(f(X)) = X$.*

(4) *For any $X \subseteq U$,*

$$f(\overline{apr}_R X) \subseteq \overline{apr}_{\hat{f}(R)} f(X).$$

(5) *If f is predecessor-consistent with respect to R, then*

$$f(\overline{apr}_R X) = \overline{apr}_{\hat{f}(R)} f(X)$$

*for any $X \subseteq U$.*

Before giving the proof of the theorem, let us briefly compare it with Theorem 3.5. In the above theorem, the assertion (1) is the same as the corresponding one in Theorem 3.5; (2) and (4) are newly added; (3), following immediately from (1) and (2), greatly improves the third assertion in Theorem 3.5, because the bijection of $f$ is much stronger than that $f$ is surjective and successor-consistent with respect to $R$. In fact, if $f$ is bijective, then $f$ is injective, surjective, predecessor-consistent, and successor-consistent, and moreover, $f^{-1}(f(X)) = X$ for any $X \subseteq U$. (5) amends the fourth assertion and significantly improves the fifth and sixth assertions in Theorem 3.5.

*Proof of Theorem 3.6.* One may refer to [16] for the proof of (1). (3) is a direct corollary of (1) and (2). Hence, we only need to verify (2), (4), and (5).

Let us start with (2). Suppose that $f$ is surjective and $f^{-1}(f(X)) = X$. For any

$$y \in \underline{apr}_{\hat{f}(R)} f(X),$$

we have that $\hat{f}(R)_s(y) \subseteq f(X)$, as

$$\underline{apr}_{\hat{f}(R)} f(X) = \{y \in V \mid \hat{f}(R)_s(y) \subseteq f(X)\}$$

by definition. In the case of $\hat{f}(R)_s(y) = \emptyset$, since $f$ is a surjective mapping, there exists $x \in U$ such that $f(x) = y$ and $R_s(x) = \emptyset \subseteq X$. Therefore,

$$x \in \underline{apr}_R X = \{x \in U \mid R_s(x) \subseteq X\},$$

and thus,

$$y = f(x) \in f(\underline{apr}_R X).$$

If $\hat{f}(R)_s(y) \neq \emptyset$, then for any $y' \in \hat{f}(R)_s(y)$, there is $(x, x') \in R$ such that $f(x) = y$ and $f(x') = y'$. To show that

$$y = f(x) \in f(\underline{apr}_R X),$$

by the previous argument it is sufficient to show that $R_s(x) \subseteq X$. By contradiction, assume that there is some $z \in R_s(x) \setminus X$, that is, $(x, z) \in R$ and $z \notin X$. Thanks to $(x, z) \in R$, we get that $(f(x), f(z)) \in \hat{f}(R)$, namely, $(y, f(z)) \in \hat{f}(R)$. As a result, $f(z) \in \hat{f}(R)_s(y) \subseteq f(X)$, which means that $z \in f^{-1}(f(X))$. As $f^{-1}(f(X)) = X$, it forces that $z \in X$, a contradiction. Consequently, we obtain that

$$\underline{apr}_{\hat{f}(R)} f(X) \subseteq f(\underline{apr}_R X),$$

which proves (2).

Let us continue proving (4). By definition,

$$\overline{apr}_R X = \{x \in U \mid R_s(x) \cap X \neq \emptyset\},$$

and thus,
$$f(\overline{apr}_R X) = \{f(x) \mid R_s(x) \cap X \neq \emptyset\}.$$

For any $y \in f(\overline{apr}_R X)$, there exists $x \in U$ satisfying that $f(x) = y$ and $R_s(x) \cap X \neq \emptyset$. Taking $x' \in R_s(x) \cap X$, we find that $f(x') \in \hat{f}(R)_s(y) \cap f(X)$, which means that $\hat{f}(R)_s(y) \cap f(X) \neq \emptyset$. Hence,
$$y = f(x) \in \overline{apr}_{\hat{f}(R)} f(X),$$
which yields that
$$f(\overline{apr}_R X) \subseteq \overline{apr}_{\hat{f}(R)} f(X),$$
as desired.

Finally, we verify (5). Suppose that $f$ is predecessor-consistent with respect to $R$. By (4), it suffices to show that
$$\overline{apr}_{\hat{f}(R)} f(X) \subseteq f(\overline{apr}_R X).$$

For any
$$y \in \overline{apr}_{\hat{f}(R)} f(X),$$
it follows by definition that $\hat{f}(R)_s(y) \cap f(X) \neq \emptyset$. Whence, there exists $y' \in \hat{f}(R)_s(y) \cap f(X)$, which means that there is some $(x, x') \in R$ such that $f(x) = y$ and $f(x') = y'$. We thus get that $x \in R_p(x')$. On the other hand, there is some $x'' \in X$ such that $f(x'') = y'$, as $y' \in f(X)$. Therefore, $f(x') = f(x'')$, which implies that $R_p(x') = R_p(x'')$ since $f$ is predecessor-consistent with respect to $R$. Consequently, $x \in R_p(x'')$, namely, $x'' \in R_s(x)$. This, together with $x'' \in X$, forces that $x'' \in R_s(x) \cap X$. Hence, $R_s(x) \cap X \neq \emptyset$, and thus, $x \in \overline{apr}_R X$, which gives that $y = f(x) \in f(\overline{apr}_R X)$. Thereby,
$$\overline{apr}_{\hat{f}(R)} f(X) \subseteq f(\overline{apr}_R X),$$
as desired. This completes the proof of the theorem. $\square$

**Remark 3.3.** For any generalized approximation space $(U, R)$ and $X \subseteq U$, one may also define the pair of lower and upper approximations using other neighborhoods (see, for example, [18, 19]). For example, employing the predecessor neighborhood, the *lower approximation* $\underline{apr}'_R X$ and *upper approximation* $\overline{apr}'_R X$ of $X$ can be defined as
$$\underline{apr}'_R X = \{x \in U \mid R_p(x) \subseteq X\} \text{ and } \overline{apr}'_R X = \{x \in U \mid R_p(x) \cap X \neq \emptyset\},$$
respectively. Based upon the newly defined approximations, there is no difficulty to develop corresponding theorem to describe the lower and upper approximations under relation mappings. We do not go into the details here.

## 4. Conclusion

In this paper, we have unified and extended type-1 and type-2 consistent functions introduced in [16] into the notion of neighborhood-consistent functions. Furthermore, we have found that type-1 consistent functions are nothing else than predecessor-consistent functions. Based on this observation, we have explored more properties of consistent functions and induced relation mappings and improve upon several deficient assertions in [16]. With the concept of neighborhood-consistent functions, the present work can be easily generalized to other approximation spaces based on different neighborhoods. Most recently, the authors have introduced predecessor-consistent and successor-consistent functions with respect to a fuzzy relation in [21] and greatly improved some characterizations of fuzzy relation mappings presented in [13]. Besides, Yang and Xu have recently introduced the concepts of $R$-open sets, $R$-closed sets, and regular sets of a generalized approximation space $(U, R)$ in [17]. It would be interesting to examine whether consistent functions and relation mappings preserve some properties of $R$-open sets, $R$-closed sets, and regular sets.

### Acknowledgements

This work was supported by the National Natural Science Foundation of China under Grants 60821001, 60873191, and 60903152. The authors would like to thank the anonymous reviewers for some helpful suggestions.

# References

[1] Gong, Z. T., Xiao, Z. Y., 2010. Communicating between information systems based on including degrees. International Journal of General Systems 39 (2), 189–206.

[2] Graymala-Busse, J., Sedelow Jr, W., 1988. On rough sets and information system homomorphism. Bull. Pol. Acad. Sci. Tech. Sci. 36 (3), 233–239.

[3] Grzymala-Busse, J., 1986. Algebraic properties of knowledge representation systems. In: Proceedings of the ACM SIGART International Symposium on Methodologies for Intelligent Systems. ACM, pp. 432–440.

[4] Li, D. Y., Ma, Y., 2000. Invariant characters of information systems under some homomorphisms. Inform. Sci. 129, 211–220.

[5] Lin, T., 1989. Neighborhood systems and approximation in database and knowledge base systems. In: Emrich, M., Phifer, M., Hadzikadic, M., Ras, Z. (Eds.), Proc. Fourth Intern. Symp. Meth. Intell. Syst. (Poster Session), Oct. 12-15, 1989, Oak Ridge National Laboratory, Charlotte, NC. pp. 75–86.

[6] Lin, T., 1997. Neighborhood systems-A qualitative theory for fuzzy and rough sets. Advances in Machine Intelligence and Soft Computing 4, 132–155.

[7] Pawlak, Z., 1982. Rough sets. Int. J. Comput. Inform. Sci. 11 (5), 341–356.

[8] Pawlak, Z., 1991. Rough Sets: Theoretical Aspects of Reasoning about Data. Kluwer Academic Publishers, Boston.

[9] Pawlak, Z., Skowron, A., 2007. Rough sets and boolean reasoning. Inform. Sci. 177 (1), 41–73.

[10] Pawlak, Z., Skowron, A., 2007. Rough sets: Some extensions. Inform. Sci. 177 (1), 28–40.

[11] Pawlak, Z., Skowron, A., 2007. Rudiments of rough sets. Inform. Sci. 177 (1), 3–27.

[12] Pedrycz, W., Vukovich, G., 2000. Granular worlds: representation and communication problems. Int. J. Intell. Syst. 15 (11), 1015–1026.

[13] Wang, C., Chen, D., Zhu, L., 2009. Homomorphisms between fuzzy information systems. Appl. Math. Lett. 22, 1045–1050.

[14] Wang, C., Wu, C., Chen, D., 2008. A systematic study on attribute reduction with rough sets based on general binary relations. Inform. Sci. 178 (9), 2237–2261.

[15] Wang, C., Wu, C., Chen, D., Du, W., 2008. Some properties of relation information systems under homomorphisms. Appl. Math. Lett. 21, 940–945.

[16] Wang, C., Wu, C., Chen, D., Hu, Q., Wu, C., 2008. Communicating between information systems. Inform. Sci. 178, 3228–3239.

[17] Yang, L., Xu, L., 2009. Algebraic aspects of generalized approximation spaces. Int. J. Approx. Reasoning 51, 151–161.

[18] Yao, Y. Y., 1998. Constructive and algebraic methods of theory of rough sets. Inform. Sci. 109, 21–47.

[19] Yao, Y. Y., 1998. Relational interpretations of neighborhood operators and rough set approximation operators. Inform. Sci. 111, 239–259.

[20] Zhai, Y., Qu, K., 2009. On characteristics of information system homomorphisms. Theory Comput. Syst. 44 (3), 414–431.

[21] Zhu, P., Wen, Q. Y., 2010. Homomorphisms between fuzzy information systems revisited. Arxiv preprint arXiv:1002.0908.

[22] Zhu, W., 2007. Generalized rough sets based on relations. Inform. Sci. 177 (22), 4997–5011.

[23] Zhu, W., 2009. Relationship between generalized rough sets based on binary relation and covering. Inform. Sci. 179, 210–225.